\pdfoutput=1

\documentclass{article}
\usepackage{nips13submit_e,times}
\usepackage{graphicx}
\usepackage{xcolor}
\usepackage{amsmath}
\usepackage{amssymb}
\usepackage{xspace}
\usepackage{multirow}
\usepackage[numbers,sort]{natbib}
\usepackage[
    pagebackref=true,
    breaklinks=true,
    colorlinks,
    bookmarks=false]{hyperref}
\usepackage[compact]{titlesec}

\def\eg{\emph{e.g.\!}\xspace}

\def\etal{\emph{et al.\!}\xspace}

\DeclareMathOperator{\Rcal}{\mathcal{R}}
\newcommand{\figref}[1]{Fig.~\ref{#1}}

\newcommand{\sref}[1]{Sect.~\ref{#1}}

\DeclareMathOperator{\indic}{\mathbf{1}}

\titlespacing{\section}{0pt}{0pt}{0pt}
\titlespacing{\subsection}{0pt}{0pt}{0pt}

\title{Deep Inside Convolutional Networks: Visualising Image Classification Models and Saliency Maps}

\author{
Karen Simonyan \\
\And
Andrea Vedaldi \\
\And
Andrew Zisserman
}

\nipsfinalcopy

\begin{document}

\maketitle

\vspace{-3.5em}
\begin{center}
Visual Geometry Group, University of Oxford\\
\verb!{karen,vedaldi,az}@robots.ox.ac.uk!
\end{center}
\vspace{1em}

\begin{abstract}
This paper addresses the visualisation of image classification models, learnt using deep Convolutional Networks (ConvNets).
We consider two visualisation techniques, based on computing the gradient of the class score with respect to the input image.
The first one generates an image, which maximises the class score~\cite{Erhan09}, thus visualising the notion of the class, captured by a ConvNet.
The second technique computes a class saliency map, specific to a given image and class. We show that such maps can be employed for weakly supervised
object segmentation using classification ConvNets.
Finally, we establish the connection between the gradient-based ConvNet visualisation methods and deconvolutional networks~\cite{Zeiler13}.
\end{abstract}

\section{Introduction}
With the deep Convolutional Networks (ConvNets)~\cite{LeCun98} now
being the architecture of choice for large-scale image
recognition~\cite{Krizhevsky12,Ciresan12}, the problem of
understanding the aspects of visual appearance, captured inside a deep
model, has become particularly relevant and is the subject of this paper.

In previous work, Erhan~\etal~\cite{Erhan09} visualised deep models by
finding an input image which maximises the neuron activity of
interest by carrying out an optimisation  using gradient ascent
in the image space.  The method was used to visualise the hidden
feature layers of unsupervised deep architectures, such as the Deep
Belief Network (DBN)~\cite{Hinton06}, and it was later employed by
\mbox{Le~\etal~\cite{Le12}} to visualise the class models, captured by
a deep unsupervised auto-encoder.  
Recently, the problem of ConvNet visualisation was addressed by
Zeiler~\etal~\cite{Zeiler13}.  For convolutional layer visualisation,
they proposed the Deconvolutional Network (DeconvNet) architecture,
which aims to approximately reconstruct the input of each layer from
its output.

In this paper, we address the visualisation of deep image
classification ConvNets, trained on the large-scale ImageNet challenge
dataset~\cite{Berg10a}.  To this end, we make the following three
contributions.  First, we demonstrate that understandable
visualisations of ConvNet classification models can be obtained using
the numerical optimisation of the input image~\cite{Erhan09}
(\sref{sec:class_model}).  Note, in our case, unlike~\cite{Erhan09},
the net is trained in a supervised manner, so we know which neuron in
the final fully-connected classification layer should be maximised to
visualise the class of interest (in the unsupervised case,
\cite{Le12} had to use a separate annotated image set
to find out the neuron responsible for a particular
class).
To the best of our knowledge, we are the
first to apply the method of~\cite{Erhan09} to the visualisation of
\mbox{ImageNet} classification ConvNets~\cite{Krizhevsky12}.  Second, we
propose a method for computing the spatial support of a given class in
a given image (image-specific class saliency map) using a single
back-propagation pass through a classification ConvNet
(\sref{sec:class_saliency}). As discussed in~\sref{sec:graph_cut},
such saliency maps can be used for weakly supervised object
localisation.  Finally, we show in~\sref{sec:comp_deconv} that the
gradient-based visualisation methods generalise the deconvolutional
network reconstruction procedure~\cite{Zeiler13}.

\paragraph{ConvNet implementation details.}
Our visualisation experiments were carried out using a single deep ConvNet, trained on the ILSVRC-2013 dataset~\cite{Berg10a}, which includes 1.2M training images, labelled into 1000 classes.
Our ConvNet is similar to that of~\cite{Krizhevsky12} and is implemented using their \verb!cuda-convnet! toolbox\footnote{\url{http://code.google.com/p/cuda-convnet/}}, although our net is less wide,
and we used additional image jittering, based on zeroing-out random parts of an image.
Our weight layer configuration is: conv64-conv256-conv256-conv256-conv256-full4096-full4096-full1000,
where convN denotes a convolutional layer with N filters, fullM -- a fully-connected layer with M outputs.
On ILSVRC-2013 validation set, the network achieves the top-1/top-5 classification error of $39.7\%/17.7\%$, which is slightly better than
$40.7\%$/$18.2\%$, reported in~\cite{Krizhevsky12} for a single ConvNet.

\begin{figure}[hp]
\centering
\includegraphics[width=\textwidth]{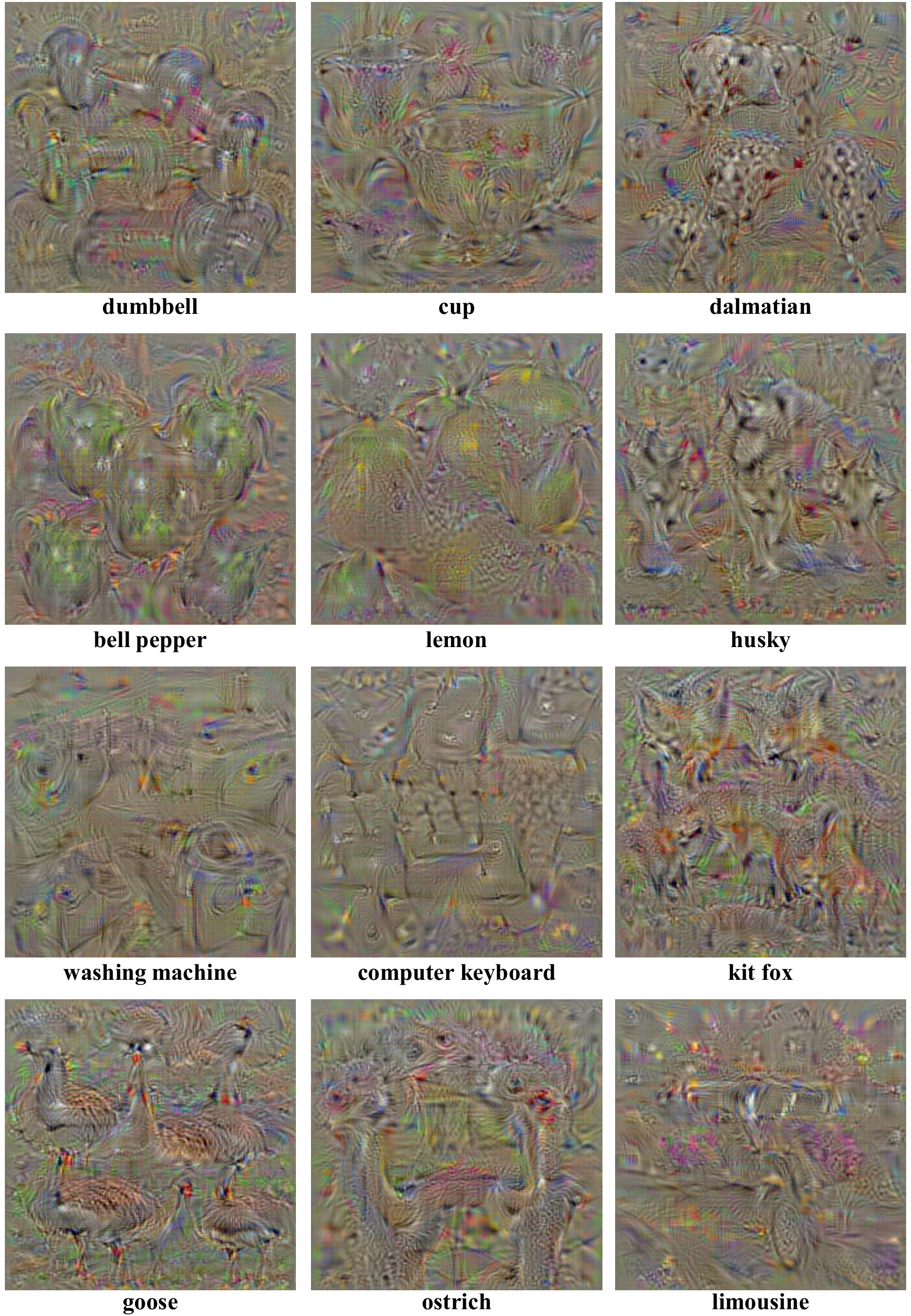}
\caption{
\textbf{Numerically computed images, illustrating the class appearance models, learnt by a ConvNet, trained on ILSVRC-2013.}
Note how different aspects of class appearance are captured in a single image.
Better viewed in colour.
}
\label{fig:class_model}
\end{figure}

\section{Class Model Visualisation}
\label{sec:class_model}
In this section we describe a technique for visualising the class models, learnt by the image classification ConvNets.
Given a learnt classification ConvNet and a class of interest, the visualisation method consists in numerically \emph{generating} an image~\cite{Erhan09}, 
which is representative of the class in terms of the ConvNet class scoring model.

More formally, let $S_c(I)$ be the score of the class $c$, computed by the classification layer of the ConvNet for an image $I$.
We would like to find an $L_2$-regularised image, such that the score $S_c$ is high:
\begin{equation}
\label{eq:class_img}
\arg\max_I S_c(I) - \lambda \|I\|_2^2,
\end{equation}
where $\lambda$ is the regularisation parameter. 
A locally-optimal $I$ can be found by the back-propagation method. The procedure is related to the ConvNet training procedure, where the back-propagation is used to optimise 
the layer weights. The difference is that in our case the optimisation is performed with respect to the input image, while the weights are fixed to those found during the training stage. 
We initialised the optimisation with the zero image (in our case, the ConvNet was trained on the zero-centred image data), and then added the training set mean image to the result.
The class model visualisations for several classes are shown in~\figref{fig:class_model}.

It should be noted that we used the (unnormalised) class scores $S_c$, rather than the class posteriors, returned by the soft-max layer: $P_c=\frac{\exp{S_c}}{\sum_c \exp{S_c}}$.
The reason is that the maximisation of the class posterior can be achieved by minimising the scores of other classes.
Therefore, we optimise $S_c$ to ensure that the optimisation concentrates only on the class in question $c$.
We also experimented with optimising the posterior $P_c$, but the results were not visually prominent, thus confirming our intuition.

\section{Image-Specific Class Saliency Visualisation}
\label{sec:class_saliency}

In this section we describe how a classification ConvNet can be queried about the spatial support of a particular class in a given image.
Given an image $I_0$, a class $c$, and a classification ConvNet with the class score function $S_c(I)$,
we would like to rank the pixels of $I_0$ based on their influence on the score $S_c(I_0)$.

We start with a motivational example. Consider the linear score model for the class $c$:
\begin{equation}
S_c(I)=w_c^T I + b_c,
\end{equation}
where the image $I$ is represented in the vectorised (one-dimensional) form, and $w_c$ and $b_c$ are respectively the weight vector and the bias
of the model. In this case, it is easy to see that the magnitude of elements of $w$ defines the importance of the corresponding pixels of $I$ for the
class $c$.

In the case of deep ConvNets, the class score $S_c(I)$ is a highly non-linear function of $I$, so the reasoning of the previous paragraph can not be immediately
applied. However, given an image $I_0$, we can approximate $S_c(I)$ with a linear function in the neighbourhood of $I_0$ by computing the first-order Taylor expansion:
\begin{equation}
S_c(I) \approx w^T I + b,
\end{equation}
where $w$ is the derivative of $S_c$ with respect to the image $I$ at the point (image) $I_0$:
\begin{equation}
\label{eq:deriv_img}
w=\left . \frac{\partial S_c}{\partial I} \right|_{I_0}.
\end{equation}

Another interpretation of computing the image-specific class saliency using the class score derivative~\eqref{eq:deriv_img}
is that the magnitude of the derivative indicates which pixels need to be changed the least to affect the class score the most.
One can expect that such pixels correspond to the object location in the image. 
We note that a similar technique has been previously applied by~\cite{Baehrens10} in the context of Bayesian classification.

\subsection{Class Saliency Extraction}
\label{sec:sal_extraction}
Given an image $I_0$ (with $m$ rows and $n$ columns) and a class $c$, the class saliency map $M \in \Rcal^{m \times n}$ is computed as follows.
First, the derivative $w$~\eqref{eq:deriv_img} is found by back-propagation. 
After that, the saliency map is obtained by rearranging the elements of the vector $w$.
In the case of a grey-scale image, the number of elements in $w$ is equal to the number of pixels in $I_0$, so the map can be computed as
$M_{ij} = |w_{h(i,j)}|$, where $h(i,j)$ is the index of the element of $w$,
corresponding to the image pixel in the $i$-th row and $j$-th column.
In the case of the multi-channel (\eg RGB) image, let us assume that the colour channel $c$ of the pixel $(i,j)$ of image $I$ corresponds to the element of $w$ with the index $h(i,j,c)$.
To derive a single class saliency value for each pixel $(i,j)$, we took the maximum magnitude of $w$ across all colour channels: $M_{ij} = \max_c |w_{h(i,j,c)}|$.

It is important to note that the saliency maps are extracted using a classification ConvNet trained on the image labels, so \emph{no additional annotation is required} (such as 
object bounding boxes or segmentation masks). The computation of the image-specific saliency map for a single class is extremely quick, since it only requires a single back-propagation pass.

\begin{figure}[hp]
\centering
\includegraphics[width=\textwidth]{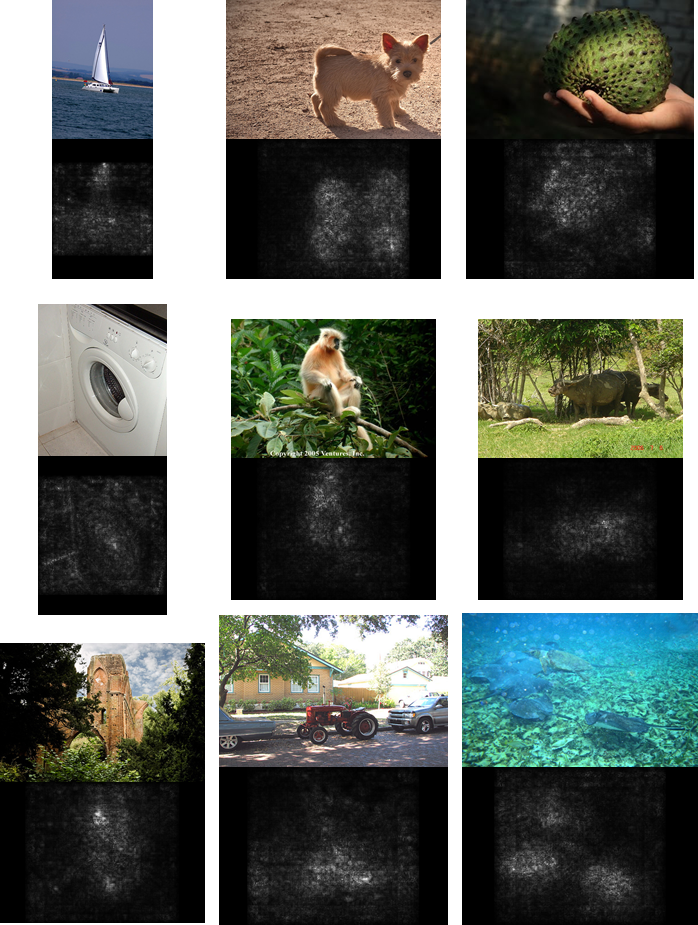}
\caption{
\textbf{Image-specific class saliency maps for the top-1 predicted class in ILSVRC-2013 test images.}
The maps were extracted using a single back-propagation pass through a classification ConvNet.
No additional annotation (except for the image labels) was used in training.
}
\label{fig:sal_map}
\end{figure}

We visualise the saliency maps for the highest-scoring class (top-1 class prediction) on randomly selected ILSVRC-2013 test set images in~\figref{fig:sal_map}.
Similarly to the ConvNet classification procedure~\cite{Krizhevsky12}, where the class predictions are computed on 10 cropped and reflected sub-images, 
we computed 10 saliency maps on the 10 sub-images, and then averaged them.

\subsection{Weakly Supervised Object Localisation}
\label{sec:graph_cut}
The weakly supervised class saliency maps (\sref{sec:sal_extraction}) encode the location of the object of the given class in the given image, and thus can be used for object localisation (in spite of being trained
on image labels only).
Here we briefly describe a simple object localisation procedure, which we used for the localisation task of the ILSVRC-2013 challenge~\cite{Simonyan13d}.

\begin{figure}[hp]
\centering
\includegraphics[width=\textwidth]{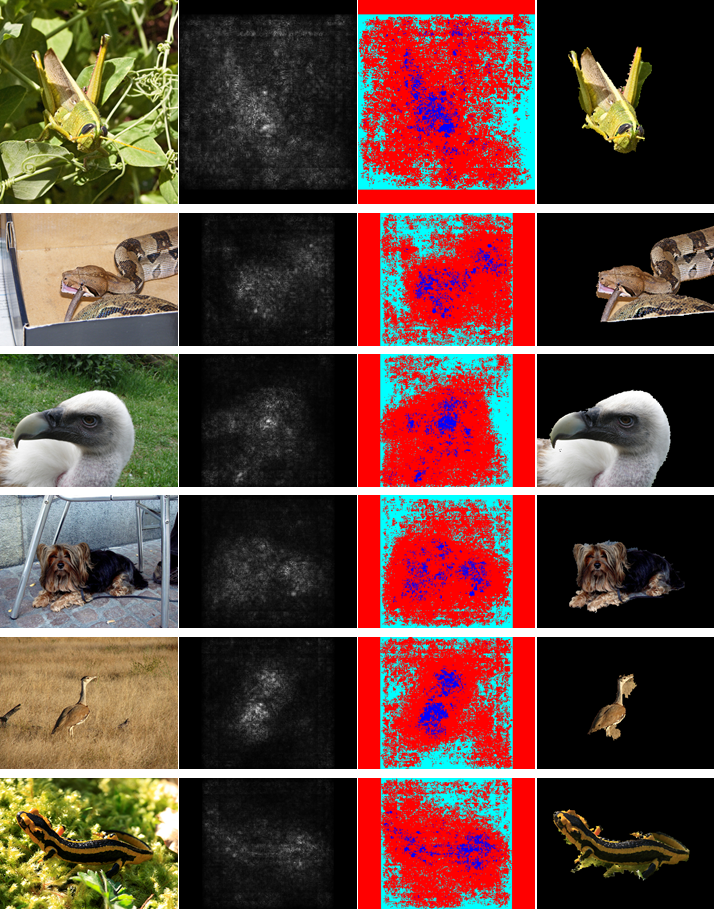}
\caption{
\textbf{Weakly supervised object segmentation using ConvNets (\sref{sec:graph_cut}).}
\emph{Left:} images from the test set of ILSVRC-2013. 
\emph{Left-middle:} the corresponding saliency maps for the top-1 predicted class. 
\emph{Right-middle:} thresholded saliency maps: blue shows the areas used to compute the foreground colour model, cyan -- background colour model, pixels shown in red are not used
for colour model estimation.
\emph{Right:} the resulting foreground segmentation masks.
}
\label{fig:seg}
\end{figure}

Given an image and the corresponding class saliency map, we compute the object segmentation mask using the GraphCut colour segmentation~\cite{Boykov01}.
The use of the colour segmentation is motivated by the fact that the saliency map might capture only the most discriminative part of an object, so saliency thresholding might not be able to
highlight the whole object. Therefore, it is important to be able to propagate the thresholded map to other parts of the object, which we aim to achieve here using the colour continuity cues.
Foreground and background colour models were set to be the Gaussian Mixture Models. 
The foreground model was estimated from the pixels with the saliency higher than a threshold, set to the $95\%$ quantile of the saliency distribution in the image;
the background model was estimated from the pixels with the saliency smaller than the $30\%$ quantile (\figref{fig:seg}, right-middle).
The GraphCut segmentation~\cite{Boykov01} was then performed using the publicly available implementation\footnote{\url{http://www.robots.ox.ac.uk/~vgg/software/iseg/}}.
Once the image pixel labelling into foreground and background is computed, the object segmentation mask is set to the largest connected component of the foreground pixels (\figref{fig:seg}, right).

We entered our object localisation method into the ILSVRC-2013 localisation challenge.
Considering that the challenge requires the object bounding boxes to be reported, we computed them as the bounding boxes of the object segmentation masks.
The procedure was repeated for each of the top-5 predicted classes.
The method achieved $46.4\%$ top-5 error on the test set of ILSVRC-2013.
It should be noted that the method is weakly supervised (unlike the challenge winner with $29.9\%$ error), and the object localisation task was not taken into account during training.
In spite of its simplicity, the method still outperformed our submission to ILSVRC-2012 challenge (which used the same dataset), which achieved $50.0\%$ localisation error using a fully-supervised
algorithm based on the part-based models~\cite{Felzenswalb08} and Fisher vector feature encoding~\cite{Perronnin10a}.

\section{Relation to Deconvolutional Networks}
\label{sec:comp_deconv}
In this section we establish the connection between the gradient-based visualisation and the \mbox{DeconvNet} architecture of~\cite{Zeiler13}.
As we show below, DeconvNet-based reconstruction of the $n$-th layer input $X_n$ is either equivalent or similar to computing the gradient
of the visualised neuron activity $f$ with respect to $X_n$, so DeconvNet effectively corresponds to the gradient back-propagation through a ConvNet.

For the convolutional layer $X_{n+1}=X_n \star K_n$, the gradient is computed as $\partial f / \partial X_n = \partial f / \partial X_{n+1} \star \widehat{K_n}$,
where $K_n$ and $\widehat{K_n}$ are the convolution kernel and its flipped version, respectively. 
The convolution with the flipped kernel exactly corresponds to computing the $n$-th layer reconstruction $R_n$ in a DeconvNet: $R_n = R_{n+1} \star \widehat{K_n}$.

For the RELU rectification layer $X_{n+1}=\max(X_n, 0)$, the sub-gradient takes the form:
$\partial f / \partial X_n = \partial f / \partial X_{n+1} \indic\left(X_n >0\right)$, where $\indic$ is the element-wise indicator function. This is slightly different from
the DeconvNet RELU reconstruction:  $R_n = R_{n+1} \indic\left(R_{n+1} >0\right)$, where the sign indicator is computed on the output reconstruction $R_{n+1}$ instead of the layer input $X_n$.

Finally, consider a max-pooling layer $X_{n+1}(p) = \max_{q \in \Omega(p)} X_n(q)$, where the element $p$ of the output feature map is computed by pooling over the corresponding spatial
neighbourhood $\Omega(p)$ of the input. The sub-gradient is computed as 
$\partial f / \partial X_n(s) = \partial f / \partial X_{n+1} (p) \indic (s = \arg\max_{q \in \Omega(p)} X_n(q))$.
Here, $\arg\max$ corresponds to the max-pooling ``switch'' in a DeconvNet. 

We can conclude that apart from the RELU layer, computing the approximate feature map reconstruction $R_n$ using a DeconvNet is equivalent to computing
the derivative $\partial f / \partial X_n$ using back-propagation, which is a part of our visualisation algorithms.
Thus, gradient-based visualisation can be seen as the generalisation of that of~\cite{Zeiler13}, since the gradient-based techniques can be applied to the visualisation of activities 
in any layer, not just a convolutional one. In particular, in this paper we visualised the class score neurons in the final fully-connected layer.

It should be noted that our class model visualisation (\sref{sec:class_model}) depicts the notion of a class, memorised by a ConvNet, and is not specific to any particular image.
At the same time, the class saliency visualisation (\sref{sec:class_saliency}) is image-specific, and in this sense is related to the image-specific convolutional layer visualisation of~\cite{Zeiler13}
(the main difference being that we visualise a neuron in a fully connected layer rather than a convolutional layer).

\section{Conclusion}
In this paper, we presented two visualisation techniques for deep classification ConvNets. 
The first generates an artificial image, which is representative of a class of interest.
The second computes an image-specific class saliency map, highlighting the areas of the given image, discriminative with respect to the given class.
We showed that such saliency map can be used to initialise GraphCut-based object segmentation without the need to train dedicated segmentation or detection models.
Finally, we demonstrated that gradient-based visualisation techniques generalise the DeconvNet reconstruction procedure~\cite{Zeiler13}.
In our future research, we are planning to incorporate the image-specific saliency maps into learning formulations in a more principled manner.

\section*{Acknowledgements}
This work was supported by ERC grant VisRec no. 228180. 
We gratefully acknowledge the support of NVIDIA Corporation with the donation of the Tesla K40 GPU used for this research.

\small
\bibliographystyle{plainnat}
\bibliography{bib/shortstrings,bib/vgg_local,bib/vgg_other,bib/new}
\end{document}